\documentclass{article}

\usepackage{microtype}
\usepackage{graphicx}
\usepackage{subfigure}
\usepackage{booktabs} 
\usepackage{amsmath}
\usepackage{float}
\usepackage{siunitx}
 \usepackage[stable]{footmisc}

\usepackage[algo2e,ruled,vlined]{algorithm2e}

\SetKwInput{KwInput}{Input}                
\SetKwInput{KwOutput}{Output}              

\usepackage{hyperref}

\usepackage[accepted]{icml2019}

\icmltitlerunning{A Hierarchical Architecture for Decision-Making in Autonomous Driving using Deep RL}

\begin{document}

\twocolumn[
\icmltitle{A Hierarchical Architecture for Sequential Decision-Making in Autonomous Driving using Deep Reinforcement Learning}



\icmlsetsymbol{equal}{*}

\begin{icmlauthorlist}
\icmlauthor{Majid Moghadam}{UCSC}
\icmlauthor{Gabriel Hugh Elkaim}{UCSC}
\end{icmlauthorlist}

\icmlaffiliation{UCSC}{Computer Science and Engineering Department, University of California, Santa Cruz, USA}

\icmlcorrespondingauthor{Majid Moghadam}{mamoghad@ucsc.edu}

\icmlkeywords{Machine Learning, Reinforcement Learning, Autonomous Vehicles, ADAS, Autonosmous Driving}

\vskip 0.3in
]



\printAffiliationsAndNotice{}  

\begin{abstract}
Tactical decision making is a critical feature for advanced driving systems, that incorporates several challenges such as complexity of the uncertain environment and reliability of the autonomous system. In this work, we develop a multi-modal architecture that includes the environmental modeling of ego surrounding and train a deep reinforcement learning (DRL) agent that yields consistent performance in stochastic highway driving scenarios. To this end, we feed the occupancy grid of the ego surrounding into the DRL agent and obtain the high-level sequential commands (i.e. lane change) to send them to lower-level controllers. We will show that dividing the autonomous driving problem into a multi-layer control architecture enables us to leverage the AI power to solve each layer separately and achieve an admissible reliability score. Comparing with end-to-end approaches, this architecture enables us to end up with a more reliable system which can be implemented in actual self-driving cars.
\end{abstract}

\section{Introduction}
\label{S:Introduction}

Advanced Driving Assistance Systems (ADAS) are developed to increase traffic safety by reducing the impact of human errors. The evolution of various levels of driving autonomy has seen a significant speed up in last years aiming to enhance comfort, safety and driving experience. For a long time, with a limited amount of technological resources, automotive stakeholders were focusing on steady-state maneuvers to achieve some sort of autonomy. However, in recent years the major focus of research in the field of autonomous driving is being directed to the transition maneuvers, most importantly tactical lane changing, required for both fully and partially autonomous driving systems \cite{ALIZADEH2019}.

Before AI, control and orientation of ground vehicles were tackled using feedback control techniques \cite{falcone2007predictive,broggi1999argo,moghadam2015actuator} that attempt to stabilize the vehicle using the information collected from sensory measurements. The controller used to generate input commands to the actuators like steering angle, accelerator, and break to perform the driving tasks. Because of the extraordinary success of the neural networks (NN) in the classification \cite{krizhevsky2012imagenet} and regression \cite{specht1991general} problems, the researchers have decided to apply learning-based approaches to the control problems. Leveraging Deep Reinforcement Learning (DRL) agents to play Atari games \cite{mnih2015human} was one of the earliest methods that gained considerable attention in the field. After the success of DRL on video game plays, researchers applied DRL on real-world problems like control and orientation of autonomous systems which have shown a great potential to become a reliable alternative for the classical control approaches.
\begin{figure}[!t]
	\centering
	\includegraphics[width=8cm]{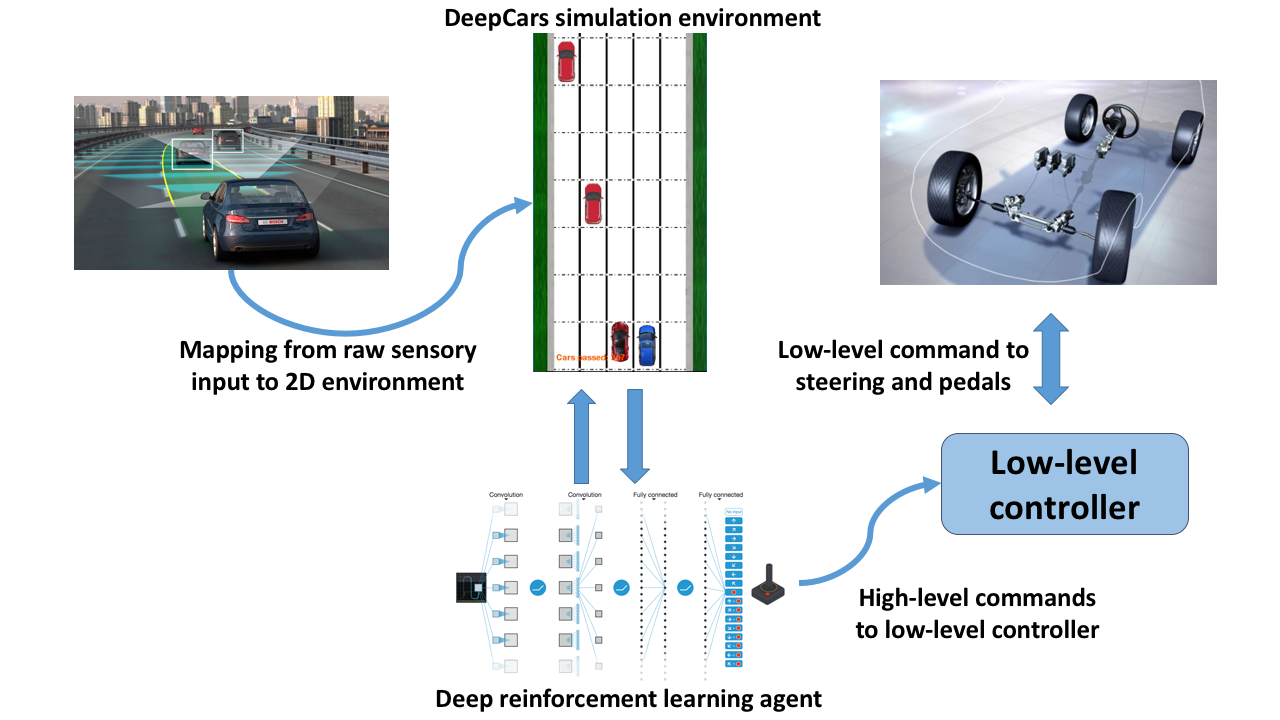}
	\caption{A sketch of our hierarchical approach for the autonomous driving problem}
	\label{f:Sketch}
\end{figure}
There has been various deep learning (DL) techniques that applied to the problem of autonomous vehicles. One approach is to use an expert knowledge of the problem and train a neural network to learn the expert policies in an end-to-end manner. This method is known as imitation learning and has been applied to both self-driving cars \cite{bojarski2016end} and unmanned aerial vehicles \cite{bicer2019vision}. Although the results are promising for training an intelligent agent that shows off comparable performance as human, such agents may never outperform the human expert as the training dataset is being annotated using solely expert demonstrations. Similar end-to-end architecture is also used to train DRL agents to drive autonomous cars in simulation environments, like TORCS \cite{koutnik2013evolving} and World Rally Championship \cite{perot2017end}. Here the term end-to-end indicates that the raw pixel information of on-board camera on the vehicles are being used as the state representation and continuous steering angle and acceleration is being calculated via DRL algorithm. End-to-end DRL approaches have the ability to outperform the human expert in various driving tasks. This can be verifies by the results published in the mentioned articles. However, the safety and reliability of such algorithms become critical problems when applied to real-world problems. In addition, the complexity of the environment in real-world applications may impeach the performance of these algorithms. In other words, combining different layers of the ADAS systems which took long to reach the current performance in an end-to-end architecture which maps the observations directly to the actions, does not seem to guarantee the similar performance in actual systems.

\subsection{Our approach}

\begin{figure}[!h]
	\centering
	\includegraphics[width=8cm]{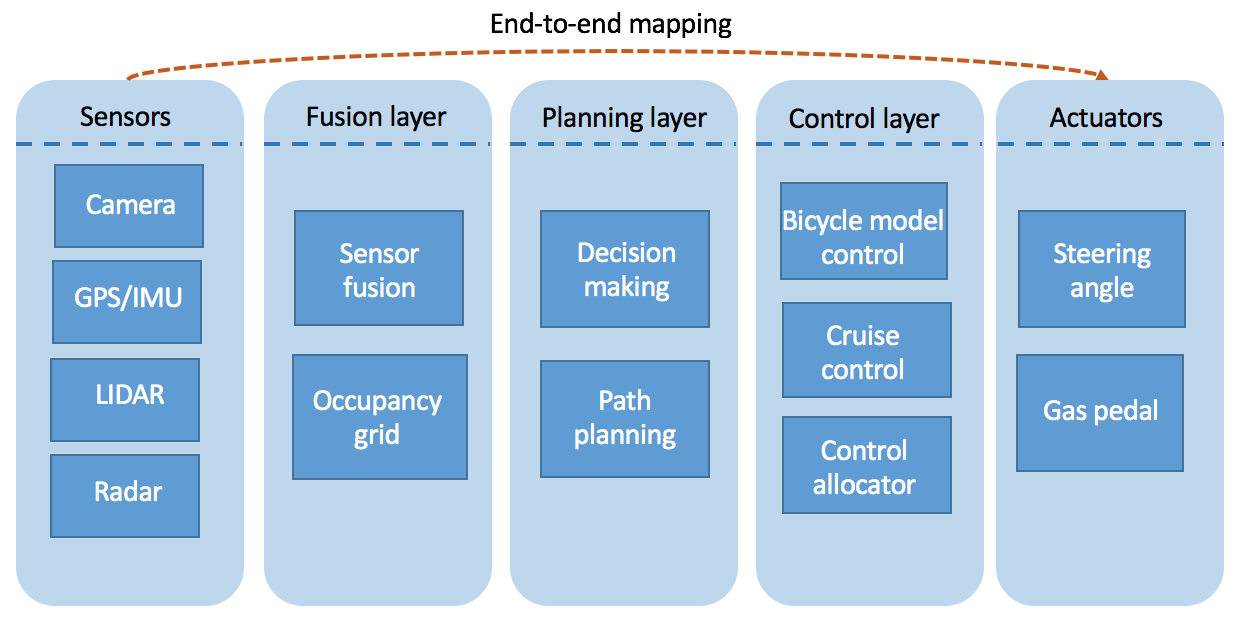}
	\caption{Hierarchical architecture of the general ADAS systems vs. end-to-end approaches}
	\label{f:ADAS}
\end{figure}

As discussed above, in most of the studies in the literature the raw sensory inputs like video frames of on-board camera or RGB-D sensors on automobile have been used to train a neural network in order to estimate the required action to control the vehicle (see Fig.\ref{f:ADAS}). Most of the studies validated the performance of their methods in a number of unreal simulations or video games. We believe that when it comes to the real-world implementation of these methods, the end-to-end learning techniques may suffer the lack of reliability to control the agent as a result of huge number of uncertainties while interacting in the actual world. To this end, we divide the problem into different control level problems and try to solve each level separately. In other words, the raw sensory inputs are going to be used to percept the environment and map it to a simpler two-dimensional world. Accordingly, the AI agent in decision making layer will generate high-level decision commands like left/right lane change and send them to lower-level control layers which deal with steering angle and gas pedal commands to stabilize the vehicle while following the generated high-level actions. This approach is summarized in Fig.\ref{f:Sketch}. Also the ADAS architecture is provided in Fig.\ref{f:ADAS}. Note that we use the occupancy grid as the environment model around the ego vehicle. We believe that, this approach is more reliable in real-world implementations than end-to-end techniques available in the literature because we do not perform a single mapping from the highest level observations with considerable number of uncertainties to the lowest level actuator control layer. This approach is much like the multi-loop control techniques in control engineering which has proven to show better performance in stabilizing complex systems \cite{slotine1991applied, moghadam2018autonomous}.

In this study, we address the problem of high-level decision making for an autonomous car using classical reinforcement learning technique known as Q-learning. We implement the $\epsilon-$greedy algorithm to the problem defined in DeepCars simulation environment which is also designed and implemented by the authors. After commenting on the performance of Q-learning algorithm, we implement the deep reinforcement learning method for the same problem. Finally we present the results for both approaches and make a comparison between them.

\subsection{Deep Q-Learning with real-time validation} \label{subsection:DQN}

The main idea in the reinforcement learning (RL) and dynamic programming (DP) is to control an agent or a process while interacting with an environment using the observing states and rewards received from the environment \cite{busoniu2017reinforcement}. DP and RL are algorithmic methods for solving decision-making problem to achieve the desired goal throughout the interaction with the world. DP methods require the model of the system's behavior, whereas RL is a model-free approach which improves the produced policy while interacting with the environment. RL uses Markov Decision Process (MDP) \cite{busoniu2017reinforcement} to mathematically formalize discrete stochastic environment. As MDP works in discrete time, states and actions in RL are usually discrete that leads to a sequential decision-making problem. Rewards also provide an informative metric of the agent's performance, and the goal is to maximize the accumulated long-term return over the track of the interaction with the environment. Although recent advances in RL \cite{lillicrap2015continuous} enables us to apply to continuous problems, here we leverage the discrete MDP to formalize the decision-making problem in autonomous driving and train RL algorithms to achieve a favorable performance.

Q-learning evaluates how good taking an action might be at a particular state through learning the action-value function $Q(s, a)$. In Q-learning a memory table $Q[s, a]$ is built to store the Q-values for all the possible combinations of states and actions. By sampling an action from the current state, the reward $R$ and the new states are found out, by which the next action $a$ that has the maximum $Q(s', a')$ from the memory table is taken. Taking an action in a particular state has a $Q-value$ which is depicted in eq. \ref{eq:target}.

\begin{equation}
\label{eq:target}
Q(s,a) = R(s, a, s') + \gamma \underset{a'}{\text{max}} Q_k (s', a')
\end{equation}

Where $s'$ and $a'$ are next state and action respectively. 
However if the combinations of state and actions are too large or states and actions are continuous, the memory and computation requirement for action-value function $Q$ will be too high. To address this issue, Deep Q-Network (DQN) \cite{mnih2015human} is utilized that approximates the action-value function $Q(s, a)$.

In this work, as shown in Algorithm 1, \cite{mnih2015human, ALIZADEH2019}, two networks, $\theta^{-}$ and $\theta$ are created and trained, one for retrieving $Q$ values and one including all updates in the training. Ultimately we synchronize $\theta^{-}$ and $\theta$ to fix the $Q-value$ targets temporarily, so it keeps the target funciotn from changing abruptly. The loss will be calculated as in eq. \ref{eq:target_network}.

\begin{equation}
\label{eq:target_network}
L_i(\theta_i) = \underset{s, a, s', r \sim D}{\mathbf{E}}\Bigg( r + \gamma \underset{a'}{max}Q(s', a'; \theta_i^{-} - Q(s, a; \theta_i)) \Bigg)^2
\end{equation}

where the transitions $(s, a, s', r)$ are retrieved from the experience replay $D$. Experience replay is used as a buffer, which we sample a mini-batch of samples from it to train the deep neural network. Since we are randomly sampling from the replay buffer, the data is closer to i.i.d and more independent of each other that in turn makes the training more stable. In addition, $\mathbf{E}$ indicates the expectation over the probability distribution.

By utilizing the experience replay and target network, the input and output of the model turns to be more stable to train and the network behaves more like a supervised learning algorithm.

We also appended the real-time validation phase to traditional DQN in order to record the best trained model during the training. For this purpose, we define two periods, by which validation phase flag is activated and lates network weights are being recorded and exploited during validation. Depending on which period, the agent's performance is evaluated for a number of episodes and the achieved mean reward is compared to the latest maximum value. This enables the agent to record the best trained model by validating on unseen scenarios. Defining two various periods with different number of episodes helps the training to be faster and record more generalized model at the same time. 

\begin{algorithm}[h]
	\label{alg:DQN}
	\DontPrintSemicolon
	\KwInput{Initialize: \\
		\qquad replay memory $D$ to capacity $M$ \\ \qquad action-value function $Q$ with random weights $\theta$ \\ \qquad target action-value function $\hat{Q}$ with wights $\theta^{-} = \theta$ }
	\KwOutput{$Q^*$}
	initialize sequence $s_1 = \{x_1\}$ \\
	initialize preprocesses sequence $\phi_1 = \phi(s_1)$\\  
	\For{t = 1, ..., T}
	{
		$a_t = \begin{cases} \text{random action} &,\text{with probability} \, \epsilon \\
		arg\underset{a}{\text{max}}Q(\phi(s_t), a; \theta) &, \text{otherwise} \end{cases}$\\
		$r_t, x_{t+1}$: apply($a_t$)\\
		$s_{t+1} = s_t, a_t, x_{t+1}$ \\
		$\phi_{t+1} = \phi(s_{t+1})$ \\
		$D \leftarrow (\phi_t, a_t, r_t, \phi_{t+1})$ \\
		$(\phi_j, a_j, r_j, \phi_{j+1}) \leftarrow random(D)$ \\
		$y_j = r_j + \gamma \underset{a'}{\text{max}} \hat{Q}(\phi_{j+1}, a'; \theta^-)$\\   
		preform a gradient descent step on $\frac{\partial \big( y_j - Q(\phi_j,a_j, ; \theta) \big)}{\partial \theta}$\\
		\If{validation\_phase}
		{
			fix network parameters $\theta$\\
			\For{$t^{\prime}$ = 1, ...,$val\_episodes$}
			{
				take greedy actions: $a = \text{max}Q(\phi(s_{t^{\prime}}), a; \theta)$\\
				record episode rewards \\
			}
			\If{mean validation reward is increased}
			{$\theta_{backup} \leftarrow \theta$}
		}
		every $C$ steps; $\hat{Q} = Q$ 
	}
	\caption{Deep Q-Network (DQN) with  experience replay and real-time validation}
\end{algorithm}

In order to improve the DQN performance, we also implemented Double DQN (DDQN) algorithm \cite{van2016deep} which uses two $Q-$networks in order to deal with the overoptimistic value estimations. Target value estimation in Double Q-learning is performed using eq. \ref{e:DoubleQ}
\begin{equation}
Q(s,a)^{DoubleQ} = R(s,a,s^\prime) + \gamma Q(s^\prime , argmax Q(s^\prime,a,\theta); \theta^\prime)\label{e:DoubleQ}
\end{equation}

where, $\theta$ and $\theta^\prime$ represent different $Q-$network parameters, one to determine the greedy policy and the other to decide on its value.
\\ \\
\section{DeepCars Simulation Environment \footnote[2]{\href{https://github.com/MajidMoghadam2006/gym-deepcars}{https://github.com/MajidMoghadam2006/gym-deepcars}} }

\begin{figure}[!h]
	\centering
	\includegraphics[width=8cm]{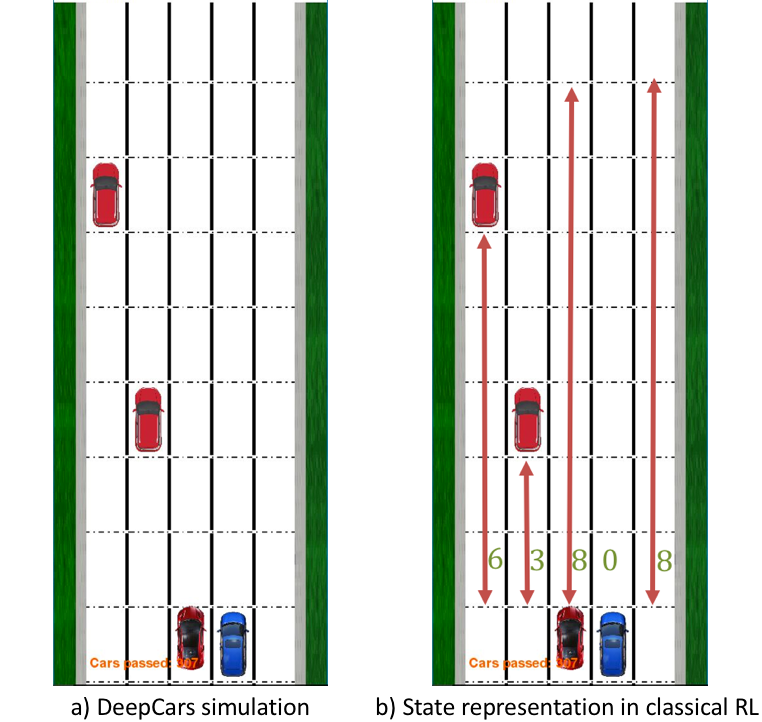}
	\caption{A screen-shot from the DeepCars 2D environment and state representation}
	\label{f:DeepCars}
\end{figure}

As discussed before, we are planning to map a 3-dimensional world into a much simpler 2D simulation environment and form the occupancy grid of the surrounding actors around the ego vehicle. To make this possible, we decided to design our own environment using the pygame (a free and open source python programming language library) to build our gaming environment. The source code is available in our GitHub repository (see footnote). We called the environment DeepCars and a screen-shot from the game screen is shown in Fig.\ref{f:DeepCars}.a.

The game receives high-level control commands as the input vector and gives the game state and reward as the output. In order to be able to formalize the problem in MDP to be used in RL algorithm, the states and actions are discrete. In fact the action space consist of three actions:
\begin{equation}
\mathcal{A} =\begin{Bmatrix}
left & stay & right
\end{Bmatrix}
\end{equation}
switch to left/right lane or stay in the same lane. The observed Markov state for each RL is explained in corresponding sections.

\section{Results}

We have implemented a tabular Q-learning, DQN, and Double-DQN (DDQN) algorithms in  DeepCars and compared the results. We will see that DDQN will demonstrate better results in terms of the performance-training speed trade-off in this problem.

\subsection{Tabular Reinforcement Learning on DeepCars}

\begin{figure}[!h]
	\centering
	\includegraphics[width=8cm]{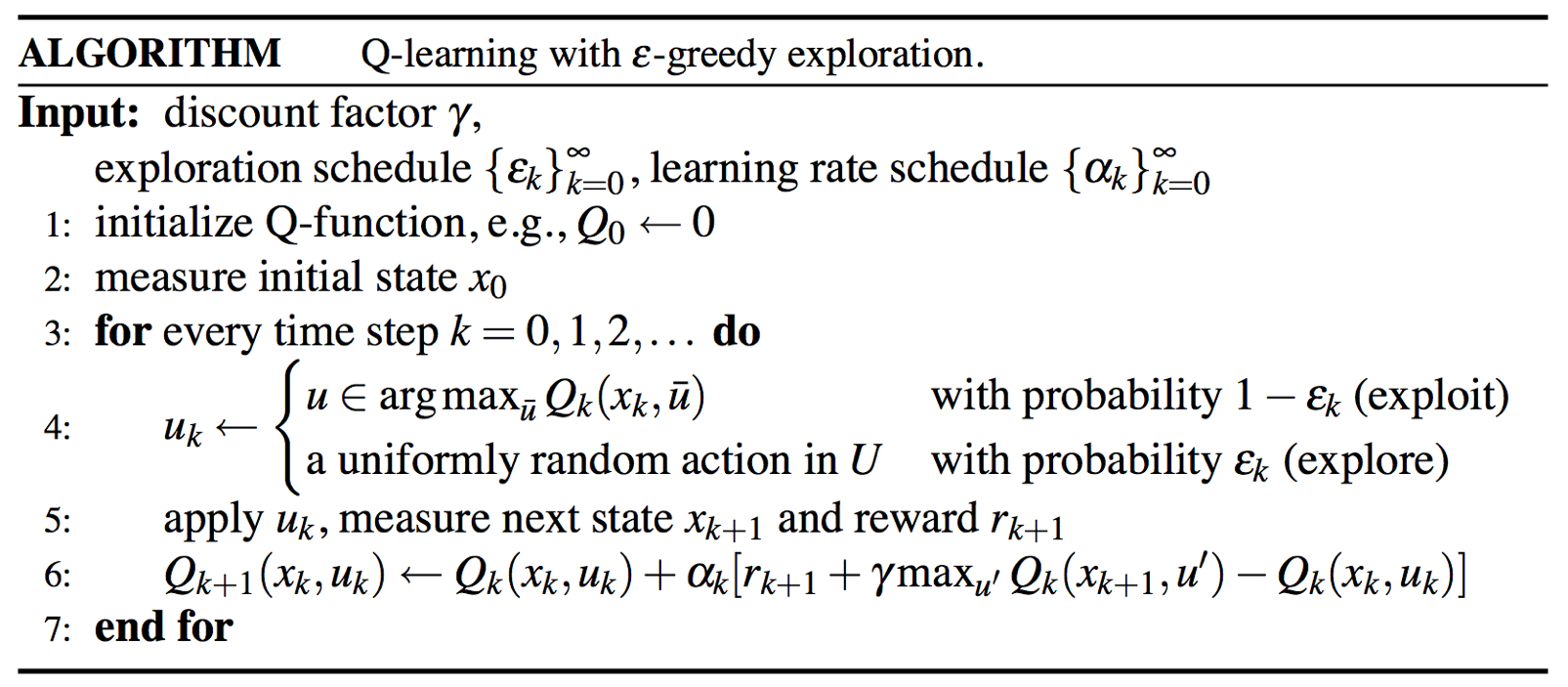}
	\caption{Tabular Q-learning algorithm \cite{busoniu2017reinforcement} where $x$ and $u$ indicate the observed state and input respectively}
	\label{f:Algorithm1}
\end{figure}

\begin{figure}[!h]
	\centering
	\includegraphics[width=8cm]{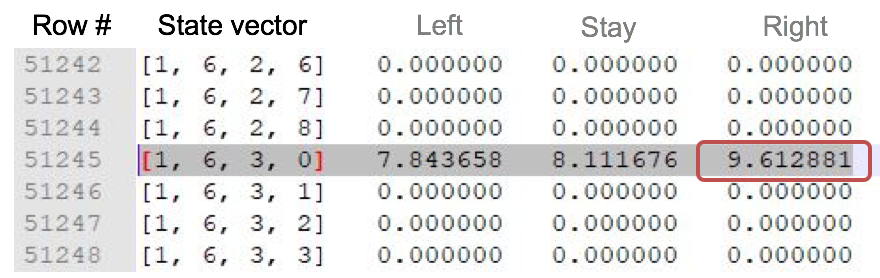}
	\caption{A crop of sparse Q-table in tabular RL (\# of lanes: 3). Red rectangle indicates the greedy optimal action where $s=\begin{bmatrix}
		1 & 6 & 3 & 0
		\end{bmatrix}$}
	\label{f:Q_table}
\end{figure}

As a starting point, we are going to implement the tabular Q-learning algorithm (Fig.\ref{f:Algorithm1}) for the control of the agent in DeepCars environment. The observed Markov state for the classical RL is defined as the following:
\begin{equation}
\label{e:RL state}
\mathcal{S} = \begin{bmatrix}
\text{ego\_lane\_ID} & x_0 & x_1 & ... & x_n
\end{bmatrix}
\end{equation}
in which the first element is the ego lane ID and $x_i$ indicates the distance to the closest car in lane $i$. Note that the lane number and grid numbers start from 0. For instance the state vector for the Fig.\ref{f:DeepCars}.b is $S = \begin{bmatrix}
2 & 6 & 3 & 8 & 0 & 8
\end{bmatrix}$.
Also note that the line-of-sight of the vehicle is 9 (from 0 to 8). This means that the furthest car that the agent can percept can be 9 grids (number of rectangles) away. This constrain in fact simulates the sensor range that are implemented on the vehicle.
\\ \\
The main objective is to train the agent to avoid making collisions with other vehicles in the environment. Thus, we define a simple reward function
\begin{equation}
\rho(s,a,s^\prime) = \left\{ 
\begin{matrix}
+1 & s^\prime\neq s_T\\
-1 & s^\prime = s_T
\end{matrix}\right.
\end{equation}
Where $s_T$ indicates the terminal state that the agent makes a collision. For the hyper-parameter tuning, we performed a simple grid search in the parameter space and found the optimal setting:
\begin{itemize}
	\item discount factor: $\gamma = 0.9$
	\item learning rate: $\alpha = 0.1$
	\item exploration: $\epsilon = 0.2$
	\item exploitation: $1-\epsilon = 0.8$
\end{itemize}
We have trained the agent for 50′000 steps. The resulting Q-table were recorded in order to be used in the evaluation phase. We evaluated the agent performance for 100'000 steps and recorded the results. In order to comment on the performance of the agent we define the accuracy as the consistent comparison metric between algorithms:
\begin{equation}
\label{E:Accuracy}
\text{Accuracy} = \frac{\text{ \# of passed cars }}{\text{ \# of all cars (passed + collided)}} \times 100
\end{equation}
The results were promising though imperfect. In fact for the training and evaluation phases, the results were as follows:
\begin{itemize}
	\item Training: Accuracy = $98.28\%$
	\item Evaluation: Accuracy = $99.14\%$
\end{itemize}
This shows that after 50′000 frames the agent has learned to control the vehicle by avoiding other vehicles. However, this is not the ideal performance, because there are a number of collisions in the test set which are not appealing. In fact our aim is to design an agent that can avoid collisions for all cases. If we take a look at the recorded Q-table after the training is finished in Fig.\ref{f:Q_table} we can see that most of the rows are still zeros which indicates a sparse Q-table that is expected in tabular Q-learning. This situation arises because not all of the state-action combinations were experienced in the training phase. And, as the agent is not using any function approximator, the similar situations cannot be approximated and collisions are unavoidable. We expect to overcome this problem by approximating Q-values using a deep neural network architecture which indicates the DQN algorithm. Note that the evaluation performance is better than training because in evaluation phase we use the best trained model. In addition, the actions are greedy in evaluation phase in contrast with $\epsilon-$greedy strategy in training phase.

\subsection{Deep Reinforcement Learning on DeepCars\footnote[3]{\href{https://github.com/MajidMoghadam2006/deepcars-reinforcement-learning}{https://github.com/MajidMoghadam2006/deepcars-reinforcement-learning}}}

 As discussed in previous section, the tabular Q-learning approach lacks the generalization property. In addition, the course of dimensionality is another problem while using Q-learning techniques. In our case, the number of Markov states ascend exponentially as the number of lanes increases. In addition, we are planning to feed the occupancy grid of the environment into the agent which makes the use of classical RL impossible for this case. This leads us to use the DQN and DDQN algorithms. The source code for DQN and DDQN algorithms are provided in our GitHub repository (see footnote). Here, we modified the original DQN algorithm and added the real-time validation phase. This helps us to perform the evaluation of the most recent model during the training and select the best trained model by far. This is possible by evaluating the latest model on a number of game episodes, say 100, and recording the agent's performance. Finally, we record the model that had the best performance at the end of training.
\\ \\
Note that in DQN the observed MDP state is the occupancy grid while in tabular Q-learning, in order to reduce the state space dimension, we defined the game state as in eq. \ref{e:RL state}. The occupancy grid incorporates the actors' position in a simple matrix with the same size as the environment grid (Fig.\ref{f:DeepCars}.a). Empty grids are filled with zeros, and ones indicate the occupied grid with an actor. The flattened version of this matrix forms the state vector. Finally, we fuse the ego lane number as a binary code of the lane ID to create the MDP state. Including ego lane ID as a binary code, enables us to have consistent binary values in the state vector, though it increases the observation space dimension.

\begin{figure}[!h]
	\centering
	\includegraphics[width=8cm]{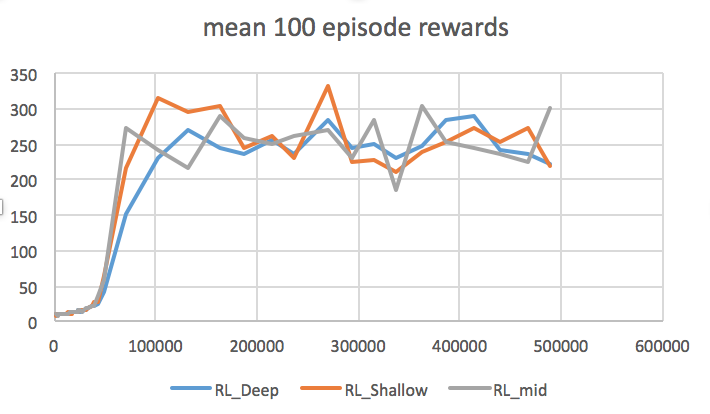}
	\caption{DQN performance for three network architectures}
	\label{f:DQN_Performance}
\end{figure}

\begin{figure}[!h]
	\centering
	\includegraphics[width=8cm]{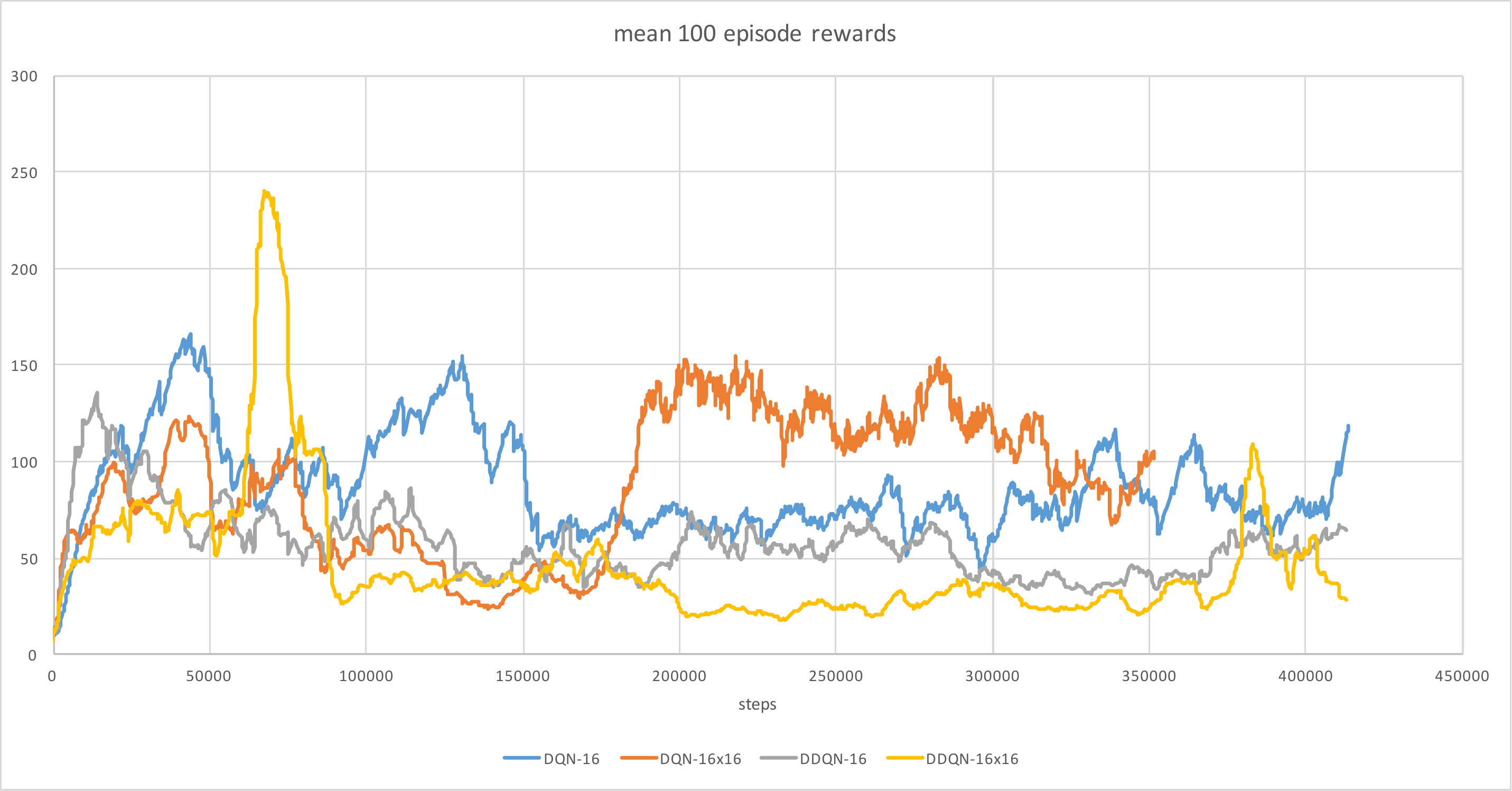}
	\caption{DQN vs. DDQN with different network architectures. Here 16x16 indicates two dense layers with 16 neurons at each.}
	\label{f:Four_agents}
\end{figure}

In addition, in order to have a smooth graph for reward values we keep track of the mean accumulated rewards every 100 episodes as the performance metric. We implemented our algorithm in a PC equipped with CPU: Intel Core i9 2.60GHz and GPU: GeForce GTX 1080 Ti. We trained the agent for 500'000 steps for three different architectures for the neural network; shallow, medium, and deep. Each differs in the number of layers of the neural network. We have used $\{32\}$ for shallow, $\{32, 64, 32\}$ for medium, and $\{64, 128, 128, 64\}$ for deep network architectures as the number of layers. The rest of the hyper-parameters were equal for all cases. The performance comparison of these networks are summarized in Fig.\ref{f:DQN_Performance}.
Obviously, the shallow network has shown a better performance comparing with two others. But the best comparison would be made in th evaluation phase where we save the trained models and evaluate them in consistent sets of evaluation episodes and keep track of the number of collisions. As shallow network promised a better performance, we use the same architecture to compare DQN and DDQN. However, we adjust the number of layers in network at the same time to search for the best option. The performance of these configurations in training phase are represented in Fig.\ref{f:Four_agents}.
Obviously, the deep version of DDQN achieved a higher reward peak value comparing with others, while shallow DDQN converged much faster than its rivals. DQN has shown a mediocre performance between these agents. Following the real-time validation policy in algorithm 1, we recorded the best model in all training phases and evaluated their performance for 100'000 steps where all of the agents achieved $100\%$ accuracy (eq. \ref{E:Accuracy}). This means that all agents are thoroughly trained and succeed to perform the evaluation phase without any collisions. This accomplishment ascertains the generalization performance of the DQN comparing with tabular Q-learning in previous section. Among deep approaches, DDQN-16 has shown the best performance-speed trade-off. Intuitively, the advanced RL algorithm (DDQN) is much faster than its predecessor (DQN) which was expected. Also, a shallow network with one hidden layer and 16 neurons seems a reasonable choice for this setting where the size of the occupancy grid as the input is $8 \times 5 = 40$. A rule of thumb to choose the number of neurons in the hidden layer is to select a number between the input and output vector sizes. These facts may justify the choice of DDQN-16 agent as the best performance-speed RL algorithm for the DeepCars simulation environment.
\\ \\
Achieving $100\%$ evaluation accuracy for the unseen scenario in training phase strengthens the assertion that it's possible to decompose the autonomous driving problem into a hierarchical architecture and leverage the AI power to solve each layer separately with an admissible reliability score.

\section{Conclusion}
In this study, we addressed the problem of autonomous driving in the highway scenario. We considered the hierarchical architecture of the ADAS systems and focused on the decision making layer, where the sequential high-level decision command are being made using a deep reinforcement learning algorithm. The agent receives the occupancy grid of the environment as the state representation and produces lane change commands in order to avoid making collisions with other vehicles. The environment is stochastic in terms of the traffic density and lane position of the other actors which challenges the agent's performance and at the same time, imitates the fusion layer output that may occur in actual test scenarios. We have shown that it's possible to leverage the provided multi-layer architecture to generate high-level commands using DRL with an acceptable reliability score.


\bibliographystyle{icml2019}
\bibliography{references}

\end{document}